\def\year{2018}\relax
\documentclass[letterpaper]{article} 
\usepackage{aaai18}  
\usepackage{times}  
\usepackage{helvet}  
\usepackage{courier}  
\usepackage{url}  
\usepackage{graphicx,subfigure}  
\usepackage{natbib}
\usepackage{hyperref}
\usepackage{url}
\pdfoutput=1
\usepackage{amssymb}
\usepackage{amsmath}
\usepackage{textcomp}
\usepackage{siunitx}
\usepackage{graphicx}
\usepackage{caption}
\usepackage{booktabs}
\usepackage{multirow}
\usepackage{tabularx}
\usepackage{algorithmic}
\usepackage{algorithm}
\nocopyright

\begin{document}
%
\title{Autonomous Extracting a Hierarchical Structure of Tasks in Reinforcement Learning and Multi-task Reinforcement Learning}
\author{ Behzad Ghazanfari \and  Matthew E. Taylor\\ School of Electrical Engineering and Computer Science\\ Washington State University \\beghazanfari@gmail.com,  taylorm@eecs.wsu.edu
}
\maketitle
\begin{abstract}
Reinforcement learning (RL), while often powerful, can suffer from slow learning speeds, particularly in high dimensional spaces. The autonomous decomposition of tasks and use of hierarchical methods hold the potential to significantly speed up learning in such domains. This paper proposes a novel practical method that can autonomously decompose tasks, by leveraging association rule mining, which discovers hidden relationship among entities in data mining. We introduce a novel method called ARM-HSTRL (Association Rule Mining to extract Hierarchical Structure of Tasks in Reinforcement Learning). It extracts temporal and structural relationships of sub-goals in RL, and multi-task RL. In particular,it finds sub-goals and relationship among them. It is shown the significant efficiency and performance of the proposed method in two main topics of RL.
\end{abstract}

\section{Introduction}

\noindent \textit{Reinforcement learning} (RL) is a common approach to learning from delayed rewards via trial and error. RL can have trouble scaling to high-dimensional state spaces due to the \textit{the curse of dimensionality} \citep{barto2003recent}. Our central thesis is that many of these issues arise from a lack of knowledge about sub-goals and the hidden relationships among them to achieve goals. Hidden relationships that depend on the tasks are a type of hierarchal knowledge: a representative structure can help define an ordering over states and sub-goals. Thus, using hierarchical structure leads a considerable improvement in the abilities of RL.

\textit{Association rule mining} (ARM) has been applied in bioinformatics \citep{bebek2007pathfinder} and large stores on market basket transactions \citep{lin2002efficient}.  For example, in big markets with millions of transactions, ARM can automatically discover items commonly being sold together \citep{tan2006introduction}. It is interesting to find out which products are bought together since the seller can put them near each other in physical supermarkets or recommend the possible products in online settings. Common correlation methods are insufficient --- it is impractical, and some of the extracted relationships are spurious. Also, some of them are unattractive because they are already known --- for more details see \cite{tan2006introduction}. ARM improves upon such simple methods by using a combination of two main measures in a proven efficient extraction strategy.

In the context of RL, sub-goals can be considered states that are correlated with successful policies to achieve goals; these states can be used to decompose the learning task \citep{digney1998learning, mcgovern2002autonomous, stolle2004automated}. Sub-goals can help an agent combat the curse of dimensionality, accelerate the agent's learning rate, and improve the quality of knowledge transfer \citep{mcgovern2002autonomous, mousavi2014automatic}. But, extracting these states automatically is completely a challenging problem --- e.g., see \cite{chiu2011subgoal}. 

In transfer learning (TL) and multi-task reinforcement learning (MTRL), different types of transferred knowledge can be used until they make learning faster and be robust to handle task differences \citep{taylor2009transfer}. We believe TL and MTRL should be done autonomously since RL is based on trial and error and is used in environments in which little is known. The proposed method extracts hierarchical structures and similar parts of different tasks autonomously. 

This paper's main contribution is to introduce and validate a method that leverages ARM to extract sub-goals and their relationships in the form of task hierarchies and implication rules in RL and MTRL.

\section{Background \& Related Work}

\noindent This section provides a brief overview of work relevant to the proposed method.

\subsection{Reinforcement Learning}

\noindent RL tasks are typically defined in the Markov Decision Process (MDP) framework as a $5-tuple$: $\langle S, A, P, R, \gamma \rangle$. In this paper, we focus on finite MDPs, where 
$S$=$\{s_{1}, \dots, s_{n}\}$ is a finite set of states, 
$A$=$\{a_{1}, \dots, a_{m}\}$ is a finite set of primitive actions, 
$P: S \times A \times S \rightarrow [0,1]$ is a one-step probabilistic state transition function, 
$R:S\times A \rightarrow \mathbb{R}$ is a reward function, 
and $\gamma \in (0,1]$ is the discount rate. 
The agent's goal is to find a policy (a mapping from states to actions), $\Pi:S \times A \rightarrow [0,1]$, that maximizes the accumulated discounted reward $R=\sum_{i=0}^{T} \gamma^i r_i$, for each state in $S$. A goal state is defined as an important, task-specific state that ends an episode once visited. 
A start state is a state from which an agent begins an episode.
%

In factored MDPs, states are described by a set of state variables. Dynamic Bayesian Networks (DBNs) can be a representative model of the transition model.

\subsection{Hierarchical Reinforcement Learning}

\noindent Temporally extended actions can take longer than 1 step and are composed of multiple primitive actions. The options framework \citep{sutton1999between} uses a Semi-Markov Decision Process (SMDP) to define temporally extended actions. An option can be defined as a 3-tuple $\langle I, \pi, \beta \rangle$, where $I \subseteq S$ is the initiation set (i.e., all states that the option can be started from them), $\Pi:S \times A \rightarrow [0,1]$ is the policy (i.e., $T-steps$ probabilistic state transition of a sequence of actions for each state in option's initiation set), and $\beta: S^{+} \rightarrow [0,1]$ is the termination condition (i.e., the probability an option terminates in a given state).


HRL is a general approach that can divide learning into several subproblems, typically leveraging options. A learning problem typically is composed of a combination of several subproblems; thus, it would be much easier to learn each of them separately and then to learn main problems based on the solution of those subproblems. Each option can correspond to learning solution for a subproblem. Many methods, including \citep{dietterich2000hierarchical,jonsson2006causal, mehta2011automatic} have shown that when a correct hierarchy is provided by an RL expert, learning can be significantly improved.

Temporally extended actions in HRL are local policies to reach local targets states. Local targets are typically categorized into two groups: bottlenecks and sub-goals. Bottlenecks are the states that provide easy access to the neighbor regions regardless of whether they are on the successful paths or they have a high gradient of reward functions such as methods which are based on a whole state transition graph \citep{mannor2004dynamic,csimcsek2004using, csimcsek2009skill}. Sub-goals are states that not only provide easy access or high reinforcement gradients, but also must be visited often \citep{mcgovern2002autonomous, stolle2004automated}. For more details see elsewhere \citep{ ghazanfari2016extracting,csimcsek2009skill}. In fact, a hierarchy of one task is learned by using temporally extended actions in the process of learning. In other words, in the options framework, temporally extended actions are often created to reach sub-goals, creating an implicit hierarchy.

If an expert has not already defined a hierarchy, several methods have been proposed to extract the task-dependent hierarchy from factored MDPs. HEX-Q \citep{hengst2003discovering} extracts a hierarchy based on an ordering of the frequency state variable changes. State variables with the highest change frequency are assigned to the lowest level of the hierarchy, and the state variable with the lowest number of changes is considered the root node. This method cannot find the relation between states variables, potentially causing learning to diverge \citep{mehta2008automatic}.

VISA \citep{jonsson2006causal} analyzes the effects between state variables by building a causal graph with a DBN. State variables that affect others are assigned to deeper levels in the hierarchy. VISA considers the effects of all actions regardless of the domain and may create unnecessary branches or unnecessary subtasks. Thus, it may not provide a reasonable hierarchy, an ``exponentially sized hierarchy” \citep{mehta2008automatic}, and performance for some problems may be poor. Such extraneous or incorrect branches may significantly reduce agent performance.

HI-MAT \citep{mehta2008automatic, mehta2011automatic} leverages a single, carefully constructed, trajectory. It leverages the factored MDP to construct a MAXQ hierarchy by constructing a DBN to build causally annotated trajectory instead of causal graphs of VISA. It is claimed the constructed hierarchy is compact and comparable to manually engineered ones. The action model is given in advance in VISA and HI-MAT, although both of them mentioned there are methods like \citep{wynkoop2008learning} to build them in advance and then use the models. 

\cite{vezhnevets2017feudal} proposed a method for extracting subgoals in  Deep RL. In\citep{bacon2017option} used different methodology ``policy gradient” instead of extracting sub-goals to create temporally extended actions. It can just handle one task and need to know the number of options in advance. It will have challenges in MTRL, and when the number of subtasks is getting increased. They can create temporally extended actions not decomposing of tasks.

\subsection{Transfer Learning and Multi-task Reinforcement Learning }

\noindent Several methods have been proposed for TL and MTRL; they can be categorized based on task differences that they can handle and the kind of transferred knowledge \citep{taylor2009transfer}. Giving hierarchical structure by a designer limits the ability of TL and MTRL since the main usages of RL is in unknown environments. The only similar work that used a task structure is  \citep{mehta2008transfer}, but a hand-made, high-level, and semantic hierarchical task structure is given by the designer to show hierarchical structure of tasks can be used for TL to handle reward function differences. ARM-HSTRL decomposes tasks in a hierarchical structure autonomously. As a result of that, it can handle more range of task differences. 

\subsection{Association Rule Mining}

\noindent The Association Rule Mining (ARM) problem is defined by $\langle ITEMSET,\ Transaction \rangle$ where $ITEMSET$ = $\{i_{1}, \dots ,i_{g}\}$ is the set of all items and $Transaction$ = $\{t_{1}, \dots ,t_{k}\}$ is the set of all transactions. Each transaction is a subset of items of $ITEMSET$. For example, Table \ref{table1} has two columns, the element set in column ``items'' of each row is a transaction. \emph{ITEMSET} = \{\emph{Bread, Movie, Beer, Coffee, Book}\}, \emph{Transaction} = $\{t_{1}, \dots, t_{7}\}$, and  $t_{1}$ = \{\emph{Bread, Movie, Beer}\}.

\begin{table}[]
	\centering
	\caption{An example of ITEMSET and Transaction } \setlength{\tabcolsep}{24pt}
	\label{table1}
	\scalebox{0.9}{
		\begin{tabular}{ll}
			Transaction ID & Items                         \\
			1              & \{Bread, Movie, Beer\}        \\
			2              & \{Coffee, Book\}              \\
			3              & \{Movie\}                     \\
			4              & \{Beer, Coffee, Movie, Book\} \\
			5              & \{Book, Coffee, Movie, Beer\} \\
			6              & \{Beer, Movie\}               \\
			7              & \{Beer, Coffee, Movie, Book\}
	\end{tabular}}
\end{table}

The relationship among states or items in the transaction set can be defined by an \textit{Association Rule}. An association rule is expressed in the form of $A\rightarrow B$, where $A$ and $B$ are disjoint itemset; $A\cap B$ = $\emptyset$. The frequency of the occurrence of $A$ and $B$ together in a data set, as disjoint items, is defined as a \textit{key factor}, also known as the \textit{support} of the association rule. The frequency of occurrence of $A$ and $B$, relative to the frequency of the occurrence of $A$, is known as the \textit{confidence}. The definition of \textit{support} and \textit{confidence} are as follows \citep{tan2006introduction}:
\[
\textit{support} (A \rightarrow B) = \frac{\sigma(A \cup B)}{N}
\]
\[
\textit{confidence} (A \rightarrow B) = \frac{\sigma(A \cup B)}{\sigma (A)}
\]

In the above equations, $\sigma()$ is a count of the number of transactions observed of the elements inside of its parenthesis and  $N$ is a count of the total number of trajectories. For example, \textit{Support} (\textit{Beer} $\rightarrow$ \textit{Movie}) equals to $5/7$ and \textit{Confidence}  (\textit{Book} $\rightarrow$ \textit{Coffee}) equals to $4/4$. In fact, \textit{support} can be used as a measure to disregard items that occur together few times, relative to the total number of trajectories. \textit{Confidence} expresses the reliability of the rule (i.e., it is the conditional probability of $B$ given $A$). There is a need to two thresholds for \textit{support} and \textit{confidence} and can be variable or fixed amounts. These two thresholds, used to disregard ``trivial'' rules, are known as \textit{minsup} and \textit{minconf} in the ARM literature \citep{tan2006introduction}. ARM algorithms typically consist of two parts: 
\begin{enumerate}
	\item Frequent Itemset Generation: all of the itemsets that satisfy the \textit{minsup} condition are extracted, i.e., frequent item sets.
	\item Rule Generation: all the rules of the previous step that satisfy \textit{minconf} rules's \textit{confidence} calculated of frequent itemset. Building upon the outputs of the Frequent Itemset Generation, this step calculates the confidence of obtained frequent itemsets and checks the eligibility of each of them by comparing their confidences with minconf threshold.
	
\end{enumerate}
As mentioned above, association rules are in the form of $A \rightarrow B$; where $A$ and $B$ are two subsets of $k$ frequent itemsets, provided $A$ and $B$ are not empty and that they satisfy the conditions of \textit{confidence} and the intersection of them is empty. 
It is impractical to enumerate all possible possibilities in a naive manner. As mentioned elsewhere \citep{ tan2006introduction}, $R$ = $3^{d}$ - $2^{d+1}$ $+$ $1$ where $R$ is the number of rules and $d$ is the number of items.
The \textit{FP-growth} algorithm has been proposed for Frequent Itemset Generation by constructing a compact data structure, a FP-tree, and based on pruning. It has been shown to work for many practical problems --- for the analysis of time complexity and more details about FP-growth algorithm see \cite{kosters2003complexity, tan2006introduction}. Thus, there are no concerns about computational complexity and practicality ARM-HSTRL --- for more details see ``Theoretical analysis and relative advantages of ARM-HSTRL'' subsection and \cite{tan2006introduction}. 
In Rule Generation, each frequent $k$-itemset has $2^{k}-2$ rules, where $k$ is the number of items of the corresponding itemset \citep{tan2006introduction}. The confidence value is calculated for each of the rules and evaluated based on \textit{minconf}. For example, if \{\textit{Beer, Movie}\} considered as a frequent itemset, its rules are \textit{Beer} $\rightarrow$ \textit{Movie} and \textit{Movie} $\rightarrow$ \textit{Beer}. \textit{Confidence}  (\textit{Beer} $\rightarrow$ \textit{Movie}) equals to $5/5$ and   \textit{Confidence}  (\textit{Movie} $\rightarrow$ \textit{Beer}) equals to $5/6$. If the assigned value of \textit{minconf} is $1$, the only association rule of the frequent item set would be  (\textit{Beer} $\rightarrow$ \textit{Movie}).

\section{ARM-HSTRL}

\noindent Key points can be extracted by observing different variations of operations and finding events that are frequently seen in successful trajectories. Similarly, ARM-HSTRL makes use of different trajectories.

ARM-HSTRL is composed of two parts (see Algorithm \ref{main}).
First, ARM extracts association rules and, second, an HST-construction converts association rules to a hierarchical structure tree. ARM is composed of two steps. First, frequent itemsets are generated. Second, the rule generation procedure applies on the output of the first step. 

\begin{algorithm} 
	\caption{ARM-HSTRL}
	\label{main}
		\begin{minipage}{1\linewidth}
			\begin{algorithmic}[1]
				\STATE \textbf{Input}$:$ Transition that is a set of successful trajectories, \textit{minsup},  \textit{minconf}	
				\STATE \textbf{Output}$:$ \textit{HST}
				
				\STATE Frequent Itemset =  FP-growth (Transition, \textit{minsup}) 
				\STATE Association Rules = Rule Generation (Frequent Itemset, \textit{minconf})
				
				\STATE HST-construction (Association Rules) //See Algorithm \ref{hstalgorithm}
			\end{algorithmic}
			
		\end{minipage}
\end{algorithm}

In ARM-HSTRL, each trajectory of visited states, $t_{k}$=$\{s_{1}, \dots, s_{h}\}$, is considered as a transaction member of Transition. All visited states in successful trajectories are in the \textit{ITEMSET}. As mentioned, sub-goals are states that are frequently visited in successful trajectories (i.e., trajectories where the agent reaches a goal state). In other words, the problem of finding sub-goals and relations among them can be seen as extracting association rules like $ \{s_{d}, \dots, s_{g} \rightarrow s_{h} \}$, where  $ \{s_{d}, \dots, s_{g}, s_{h} \}$ are sub-goal states. Subgoals are almost common for different tasks. Even for some different tasks, some key subtasks are similar; when one goes to university or supermarket or restaurant, dressing, driving, and parking car are common. In summary, ARM-HSTRL in HRL is able to analyze successful trajectories that are generated come from different start states and goals states.

The FP-growth algorithm is used for Frequent Itemset Generation. If \textit{minsup} is assigned to one, sub-goals must be visited in each trajectory of each Transaction. The maximum value of minsup is one. If the value of \textit{minsup} is too small, the performance of FP-growth decreases because it may provide some false-positive itemsets for the evaluation of Rule Generation. Some RL domains have multiple types of successful trajectories and may have different sub-goals. Thus, the value of \textit{minsup} should be set to handle such conditions. However, if \textit{minsup} is assigned to a value smaller than one, the extracted hierarchical structure would have more sub-goals, and they may be unnecessary. Then, the Rule Generation procedure, as described in ``Background \& Related Work, ''  is called on the output of FP-growth, frequent itemset. 

Recall that a confidence value is the conditional probability of the occurrence of a consequent of a rule when the premise of it is seen. Also, confidences of rules are in the form of $ {s_{f}, \dots, s_{j} \rightarrow s_{m}}$. Confidences are calculated and compared with \textit{minconf}. In addition, the confidence value of each association rule can be used as a priority score to choose among corresponding temporally extended actions of association rules.

Each extracted association rule is a set of sub-goals. It is needed to extract different possible sequences of them for HST construction. In fact, the combination of HST and ARM is a sequential association rule mining procedure.
The value of $ t$, time, of each sub-goal in each trajectory can be compared to create a sequence of seeing sub-goals.
Each sequence shows the relationship among sub-goals in a flat manner of one association rule. For example, there are two trajectories of four sub-goals ${a, b, c \rightarrow d}$ and ${b, a, c \rightarrow d}$.  $t$'s values of $a$ and $b$  are $\{1 , 2\}$ and   $\{2 , 1\}$ correspondingly  in the trajectories. If the frequency of those orders is same, it shows the order of visiting $a$ and $b$ is not important to achieve the consequent. The order of each trajectory is like a local view since different sequences to achieve goals can exist. The values of ordering of each sub-goal in all trajectories can form a range; those numbers show different possibilities of ordering subtasks. By ordering the ranges and making branches, HST construction makes a general plan from all of the possible paths.

Algorithm \ref{hstalgorithm}, HST-construction, that makes the hierarchical structure of tasks. Each rule is in form of $AR_{i}= s_{ti}, \ldots, s_{(t+n)i} \rightarrow s_{(t+n+1)i}$. $\{ s_{ti}, \ldots, s_{(t+n)i} \}$ are the sequence of sub-goals of the $AR_{i}$. $Len_{i}$ shows the number of items in $AR_{i}$,  the number of elements of the premise of the $AR_{i}$ is $n+1$ and the number of element of the consequence of each $AR$ is 1; thus, the $Len_{i}$ is $n+2$.  $AR_{i,j}$ is the $j$th element from the end of $AR_{i,j}$. For example, $AR_{i,2}$ is $s_{t+n}$ and $AR_{i,len_{i}}$ is $s_{t}$. $NumRules$ is the number of association rules.

\begin{figure}	
	\centering
	\includegraphics[width=60mm, height=1.5in]{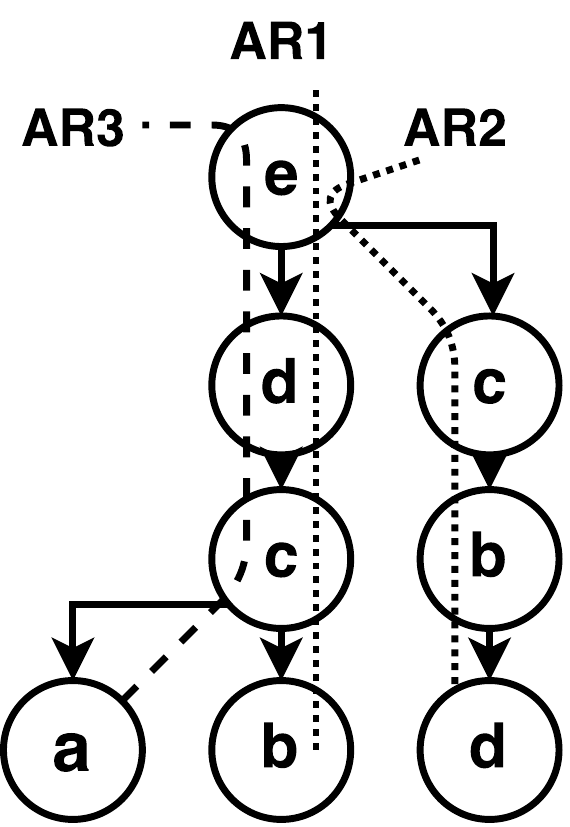}	
	\caption{ An example of a HST-construction.}
	\label{fig:HSTexample1}
\end{figure}

For example, consider  $AR_{1} = bcde$,  $AR_{2} = dbce$, $AR_{3}= acde$ (see Figure \ref{fig:HSTexample1}). First, construct the tree with the reverse of $AR_{1}$, creating one branch with values $edcb$. Then, the reverse of $AR_{2} $ is added to tree, making a new branch from $c$ since $AR_{2,2}=c$  cannot be matched in the tree from that point. Thus, a new branch from $e$ is created and the remaining values of $AR_{2}$ assigned in that. Finally, the reverse of $AR_{3}$ is added to the tree. The mismatch happens in $AR_{1,4}$ and thus a new branch is created at node $c$.

The HST helps an agent to choose temporally extended actions correctly. There is another way to extract a hierarchical structure based on sub-goals --- the extracted order is eliminated and the elements of association rules considered as separate entities like the methods that just can extract sub goals and bottlenecks. Then, the hierarchical structures can be learned by adding corresponding temporally extended actions of extracted sub-goals in the learning phase of RL.

\begin{algorithm} 
	\caption{HST-construction}
	\label{hstalgorithm}
		\begin{minipage}{1\linewidth}
			\begin{algorithmic}[1]
				\STATE \textbf{Input}$:$ \textit{AR-set} is the set of association rules. \textit{AR-set} = $\{AR_{1}, \ldots ,AR_{NumRules}\}$
				
				\STATE \textbf{Output}$:$ \textit{HST}
				\STATE \ 
				\STATE Construct a tree, $T$, with one node that is the root node, $R$. 
				\FOR { $i=1:NumRules$  } 
				\STATE Parent-Node=$R$\
				
				\FOR {$j=1:Len_{i}$} 
				\STATE $t=1$\
				\STATE $FlagM=0$\
				\REPEAT \STATE $num$ shows the number of children of the Parent-Node\
				\STATE $PN_{t}$ shows the $t_{th}$ child of the Parent-Node\ 
				\IF {$AR_{ij}==PN_{t}$}
				\STATE Parent-Node=$PN_{t}$\
				\STATE	$FlagM=1$\
				\ENDIF
				\STATE ${t++}$
				\UNTIL{$t<=num$ and $FlagM==0$}	
				
				\IF{$FlagM==0$}
				\STATE create a new child Node in the Parent-Node: $PN_{num+1}=AR_{ij} $
				\STATE Parent-Node=$PN_{num+1}$
				\ENDIF				
				\ENDFOR
				\ENDFOR

			\end{algorithmic}
		\end{minipage}
\end{algorithm}

\subsection {ARM-HSTRL in MTRL}

\noindent This subsection is presented based on the terms and definition of \cite{taylor2009transfer}.  ARM-HSTRL constructs HST of tasks that their start and goal states of each run is different in RL. But, ARM-HSTRL in MTRL can be more effective. Partial policies or options and structure can be used as transferred knowledge. In other words, it could handle other types of task differences. In ARM-HSTRL, source task selection is done based on making a library of learned tasks. They are checked, evaluated, and the most compatible and highest expectable reward one is selected. The allowed learners are hierarchical approaches (for more details see \citep{taylor2009transfer}).

The learned structures and sub-goals help agent to have a rational and quick evaluation. Checking the structure has a low cost and less prone to error, and it can infer the correct MDP and help to prioritize tasks. ARM-HSTRL can prioritize the possible tasks based on two procedures. 1) it can use the confidence of ARM to have approximations about possibilities of occurrence of each task. 2) ARM-HSTRL is like an attentional function; it extracts important states that can provide knowledge with low cost about states depended on tasks. 3) The structure of each task in the form of sub-goals is like a signature of that task. The HST is like a manual that shows the agent how to follow complicate operations. Also, it separates tasks of each other by using the sequence of sub-goals of a task as the signature of that task. 

\subsection {Theoretical analysis and relative advantages of ARM-HSTRL}
\label{Advantageous}

\noindent As mentioned in \cite{tan2006introduction}, ``the size of a FP-tree typically is smaller than the size of the uncompressed data,'' and in the worst-case scenario, the size of a FP-tree is effectively equal to the size of the data. The performance of the FP-growth algorithm is related to the compaction factor of the trajectories and the value of \textit{minsup}. In the worst-case, support values of all combination of items are bigger than \textit{minsup}, and $2^{d+1}$ itemsets will be generated, where $d$ is the number of items. However, ARM-HSTRL is looking for sub-goals, and the number of sub-goals in an RL task is much less than the size of state space. Thus, using the FP-growth algorithm is efficient and practical in ARM-HSTRL when the state space is large, and the number of sub-goals is relatively low.

ARM is proposed to work in real usages in which state space is large and sparse. If the state space is small or the successful trajectories have many similarities to each other, many states will be visited frequently, and ARM detects all of them as sub-goals. Clearly, the concept of sub-goals becomes meaningless in such conditions. Another possible scenario happens for adjacent states around sub-goals that might be visited frequently. Under both conditions, one efficient solution is clustering adjacent sub-goals as one entity and creating one corresponding temporally extended action for that entity. The $t$ of each state in each trajectory is saved; they are used in HST for possible orderings of sub-goals. They can be used to find close sub-goals and cluster them. 

If a hierarchical policy is recursively optimal, the hierarchy is a hierarchically optimal policy by definition \citep{dietterich2000hierarchical}. In the following, a sketch proof is explained for converging ARM-HSTRL to a hierarchically optimal policy -- it is similar to other methods that work in the options framework. As mentioned, subtasks are defined recursively in the form of temporally extended actions in the options framework. Learning in the options framework is proven \citep{sutton1999between} to lead to optimal policies because an agent can always revert to primitive actions. Thus, ARM-HSTRL is recursively optimal, and resulting policies are also hierarchical optimal policies. 

Sub-goal extraction methods \citep{mcgovern2002autonomous, stolle2004automated} just can find sub-goals but not hierarchical structure of tasks, they typically calculate their measures among paths of the agent or shortest paths between nodes of graph \citep{ghazanfari2016extracting}. Their performances typically become worse in more severe forms as state space becomes large, and also when the number of actions to reach goals states increase. ARM-HSTRL is based on FP-growth algorithm that is proven practical from time complexity for real usages, and it considers all paths together one time. The performance of ARM-HSTRL is independent of the number of actions; it means it can scale in continuous action space.

If the proposed method is compared theoretically with the recent HI-MAT algorithm, several differences can be highlighted. DBNs must be obtained for each state variable before HI-MAT can be run, which may be time-consuming and difficult to perfect. Also, HI-MAT needs preprocessing to find DBN-closure for each state variable value and related state variables of its casual links recursively, 
and DBN-closure for rewards --- for more details see \cite{mehta2008automatic}. 
In addition, removing unsuccessful and redundant actions cycles is necessary. Since HI-MAT works on a single successful trajectory, it should be generalized by using another function (i.e., \textit{action generalization}). HI-MAT cannot be generalized from many different starting places in a few terminal states (i.e., it does not have the \textit{funnel} property \citep{mehta2008automatic}). In many RL settings, there are several optimal or near-optimal trajectories, which will not be represented in HI-MAT. 

In contrast, our proposed method does not require a single, carefully formed, trajectory, and it can efficiently handle the funnel property of subtasks. The proposed method can work in both MDPs and factored MDPs, while the methods mentioned above (i.e., HEX-Q, VISA, and HI-MAT)  can only work in factored MDPs. Our method does not need the action model in advance or a separate phase of learning to obtain the required data for extracting hierarchy. The proposed method can easily be extended to factored MDPs by considering the value of each state variables as an item.

\begin{figure}	
	\centering
	\includegraphics[width=40mm, height=2in]{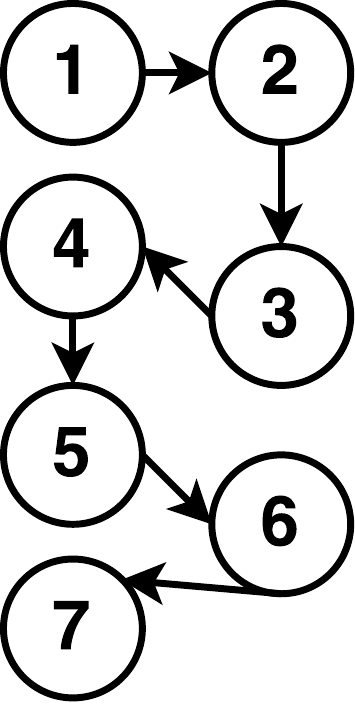}	
	\caption{ The first task hierarchy of the first testbed, Figure \ref{fig:testbed1}, for  experiment 1.}
	\label{fig:HST_test1_hi}
\end{figure}

\begin{figure}	
	\centering
	\includegraphics[width=60mm, height=2in]{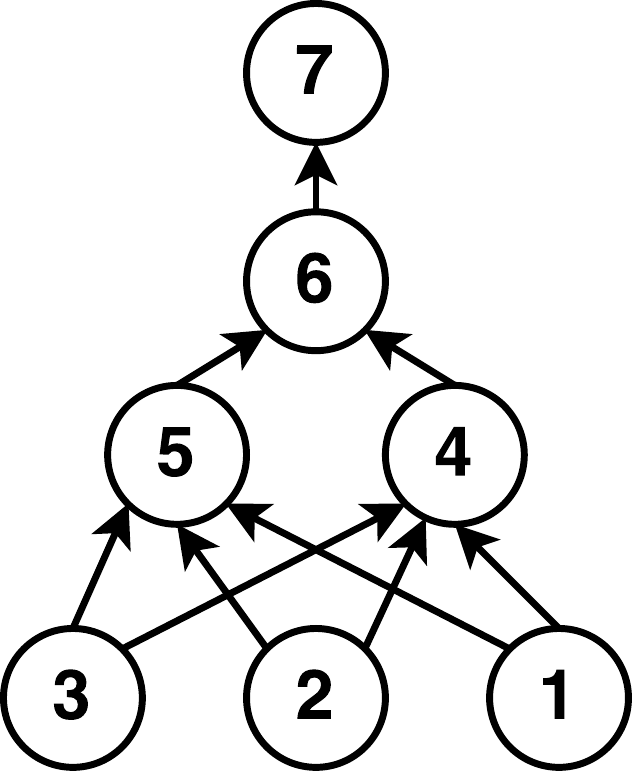}	
	\caption{ The second task hierarchy of the first testbed, Figure \ref{fig:testbed1}, for experiment 2.}
	\label{fig:HST_test2_hi}
\end{figure}

\begin{figure}	
	\centering
	\includegraphics[width=40mm, height=2in]{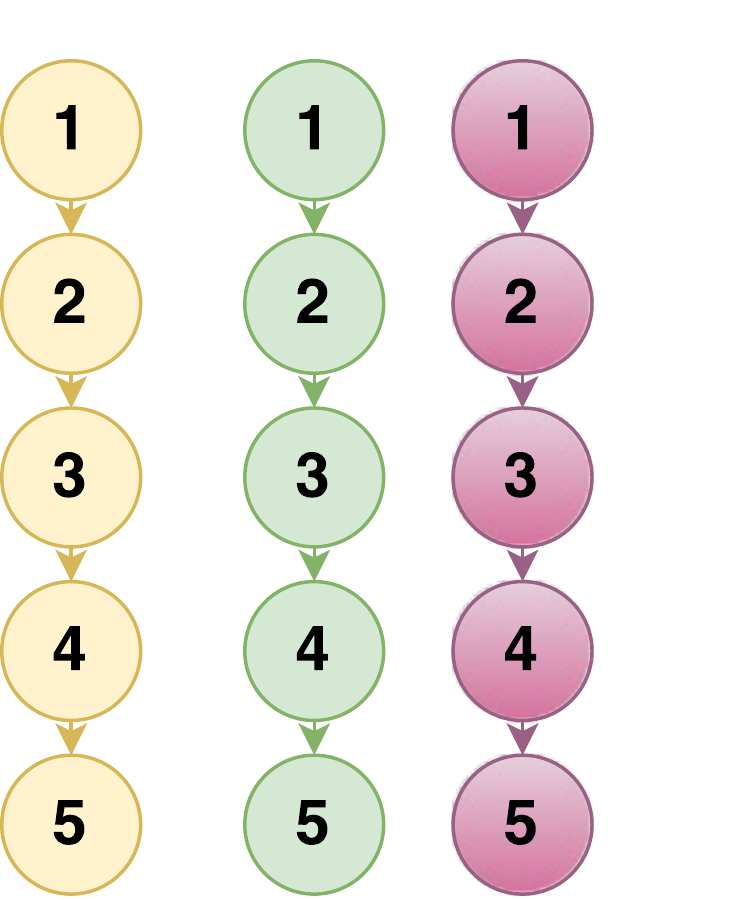}	
	\caption{ The task hierarchy of the second testbed, Figure \ref{fig:testbed_three}, for experiment 3.}
	\label{fig:HST_test3}
\end{figure}

\begin{figure}	
	\centering
	\includegraphics[width=80mm, height=2.5in]{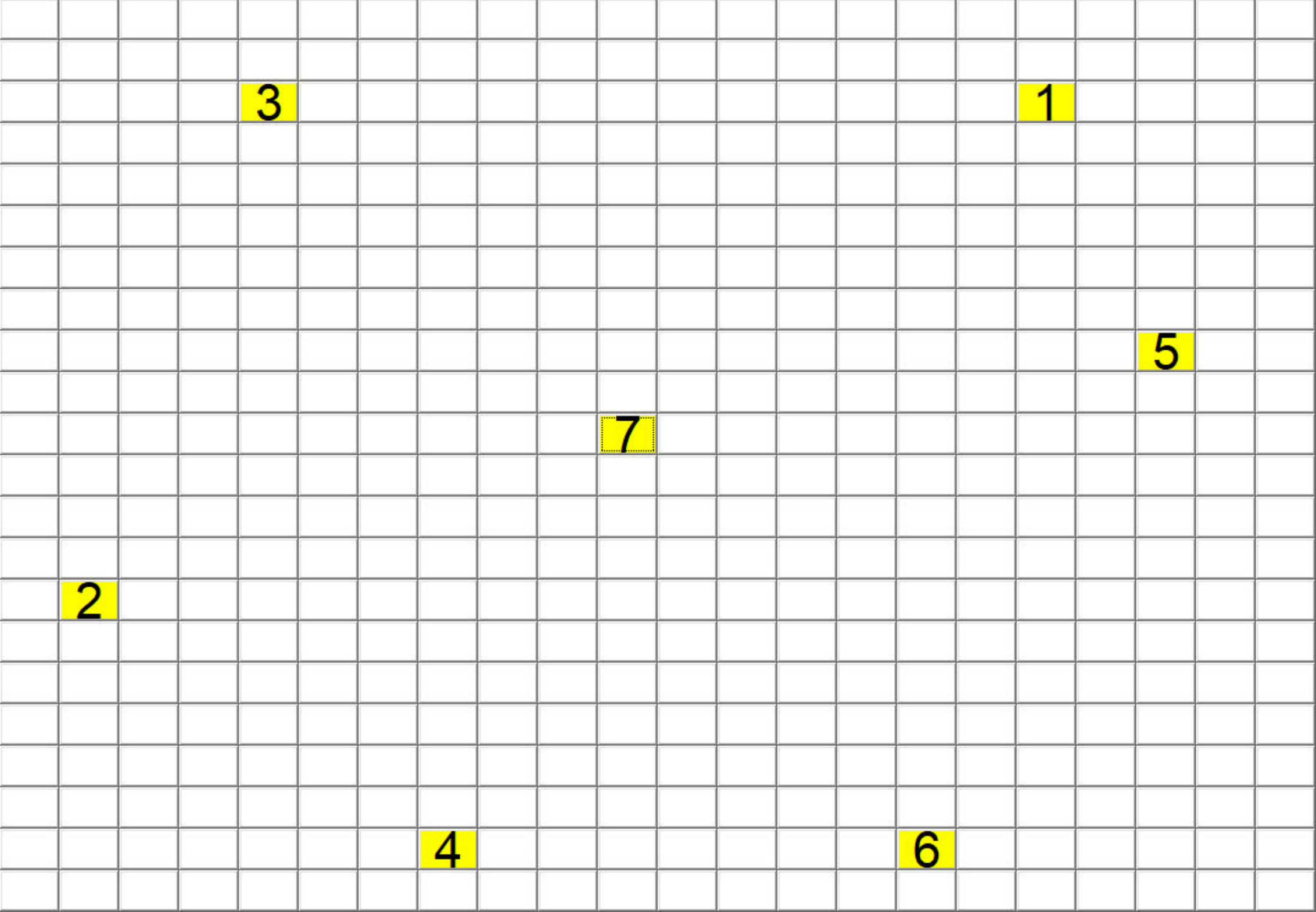}	
	\caption{ The first testbed: the size of the maze is $22\times22$ and it has 7 sub-goals. Subgoals are colored with yellow}
	\label{fig:testbed1}
\end{figure}

\begin{figure}	
	\centering
	\includegraphics[width=85mm, height=2.5in]{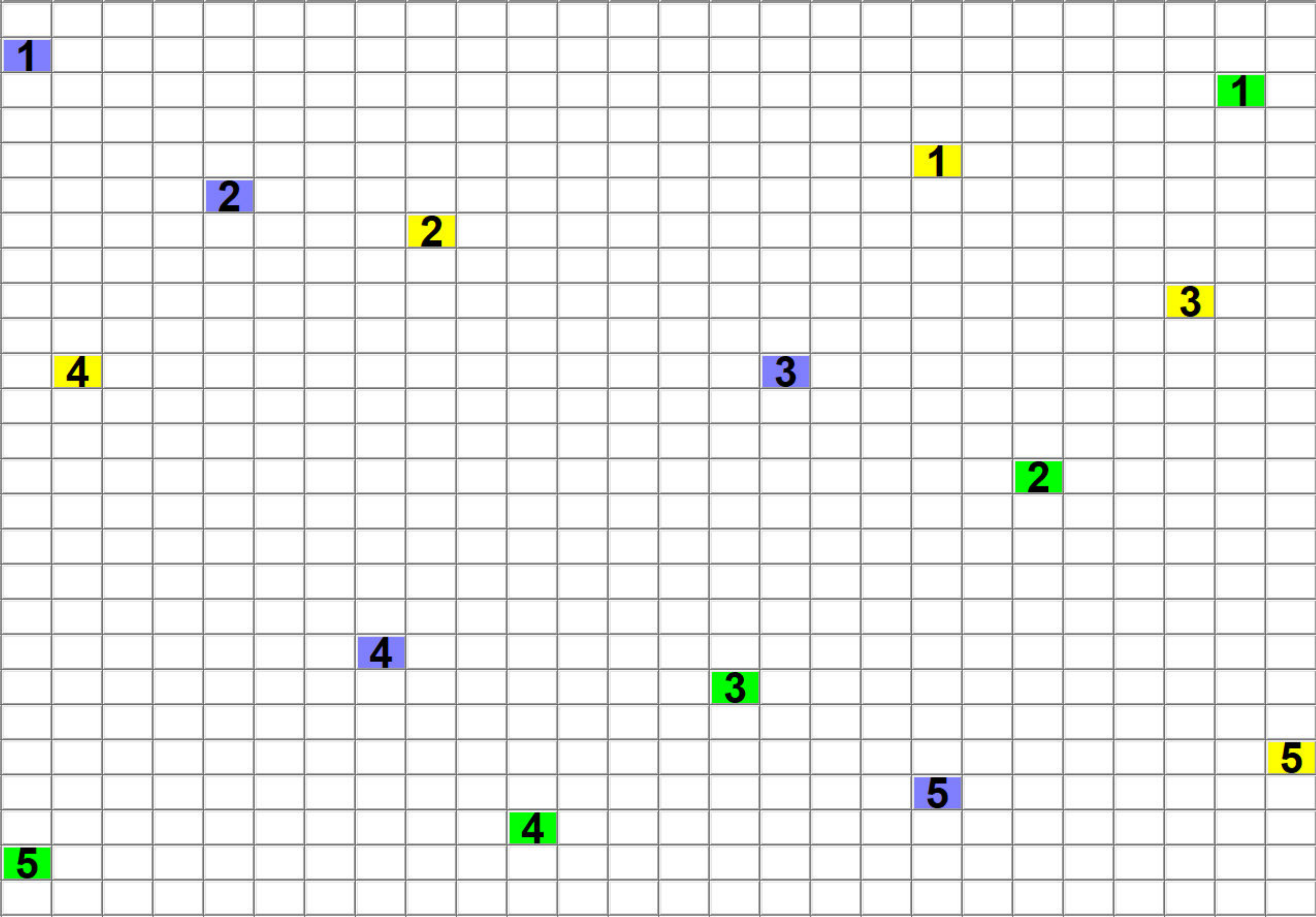}	
	\caption{ The second testbed: the size of the maze is $26\times26$. Each task has 5 sub-goals and is shown with a different color.}
	\label{fig:testbed_three}
\end{figure}

\section{Experimental Results}

\noindent In this section, three experimental results are presented on two testbeds, Figure \ref{fig:testbed1} and Figure \ref{fig:testbed_three}, for ARM-HSTRL. The agent has 5 actions, $press-key$ and 4 movement primitive actions. The $press-key$ does not change the place of the agent. The agent can move with its primitive actions in four directions: $up$, $right$, $down$, $left$. If there is a wall in the way, the agent stays in its current state. In all of the experiments, if the agent does the $press-key$ in a wrong place, it receives the reward of $0$ in the sub-goal places and the reward of $-10$ in other states. The reward for other actions is $-1$. The agent movement with probability $0.8$ is according to intended action and is randomly in one of the directions with probability $0.2$. The discount factor was set to $\gamma$= $0.9$. 

In constructing HST, $10$ start and goal places are chosen randomly. For each of them, the agent starts to learning with a learning mechanism like Q-learning; the learning is finished after 5000 episodes. They are ordered based on the accumulated reward, and the best five ones of them are selected. They are given to the ARM-HSTRL and HST will produce a hierarchal structure of tasks. Each node of the hierarchy is a sub-goal, and corresponding temporally extended actions are calculated \citep{mcgovern2002autonomous}. Now, the temporally extended actions are added to current primitive actions of the agent and HST helps agent to choose temporally extended actions along with primitive actions in the phase of learning.  Temporally extended actions are composed of primitive actions. If they considered like primitive actions, the number of steps to reach a goal is equal to the number of action selection call. 

\textbf{HRL}: ARM-HSTRL is evaluated in HRL on Figure \ref{fig:testbed1} for two different hierarchical structure of tasks, experiment 1 and experiment 2. In them, for comparison between Q-learning and ARM-HSTRL, there are 10 runs as in each of them a start and a goal state is chosen randomly, the maximum number of actions for each episode is 4000, and the total number of episodes is 8000.
In experiment 1, Figure \ref{fig:HST_test1_hi}, the task hierarchy has 7 levels -- it has $(484\times(7+1))=3872$ states. If the agent enters in sub-goals states in the following order $1,2,3,4,5,6 $ and $7$ and do the $press-key$ action in each of them, and then enters in the goal state of the run and do $press-key$ in that too, the agent receives a reward of $+10$, and the episode will be finished. The value of \textit{minsup} is 0.9 and the value of \textit{minconf}  is 0.9.

In experiment 2, Figure \ref{fig:HST_test2_hi}, the task hierarchy has 3 levels, but more complicated structure- it  has $(484\times(3+1))=1936$ states. If the agent enters in one of the sub-goals states from the leaves of tree $1$ or $2$  or $3$, then enters in one of their parent $4$ or $5$, then in $6$ and $7$ in order and does the $press-key$ action in each of them, and then enters in the goal state of the run and do $press-key$, the agent receives a reward of $+10$ and the episode will be finished. The value of \textit{minsup} is 0.3 and the value of \textit{minconf} is 0.9.

\begin{figure}	
	\centering
	\includegraphics[width=80mm, height=2in]{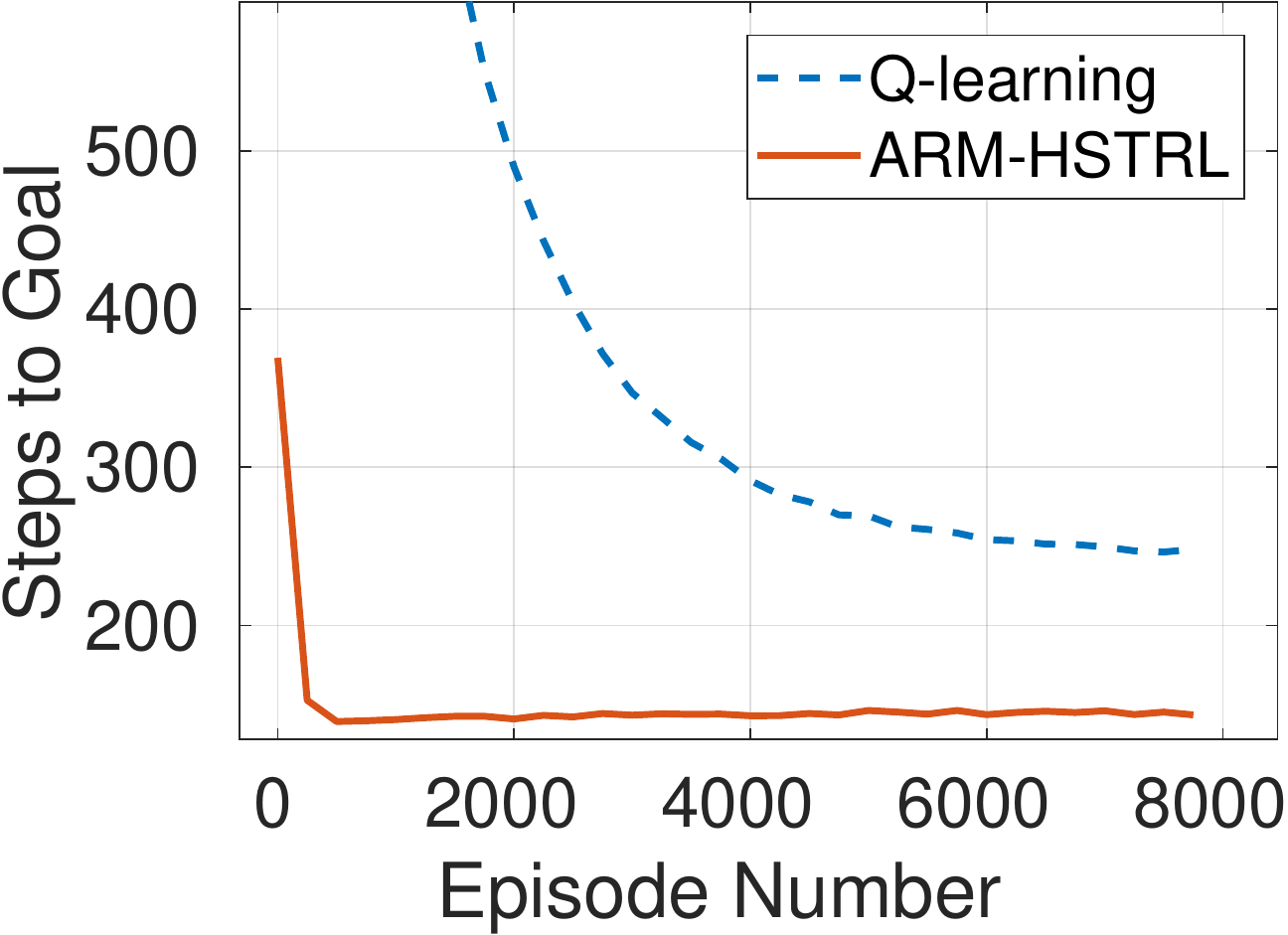}	
	\caption{ Represents the number of steps along episodes. The comparison is between Q-learning and ARM-HSTRL in experiment 1, Figure \ref{fig:HST_test1_hi},  of the first testbed, Figure \ref{fig:testbed1}.}
	\label{fig:maze_action_1_z}
\end{figure}

\begin{figure}	
	\centering
	\includegraphics[width=80mm, height=2in]{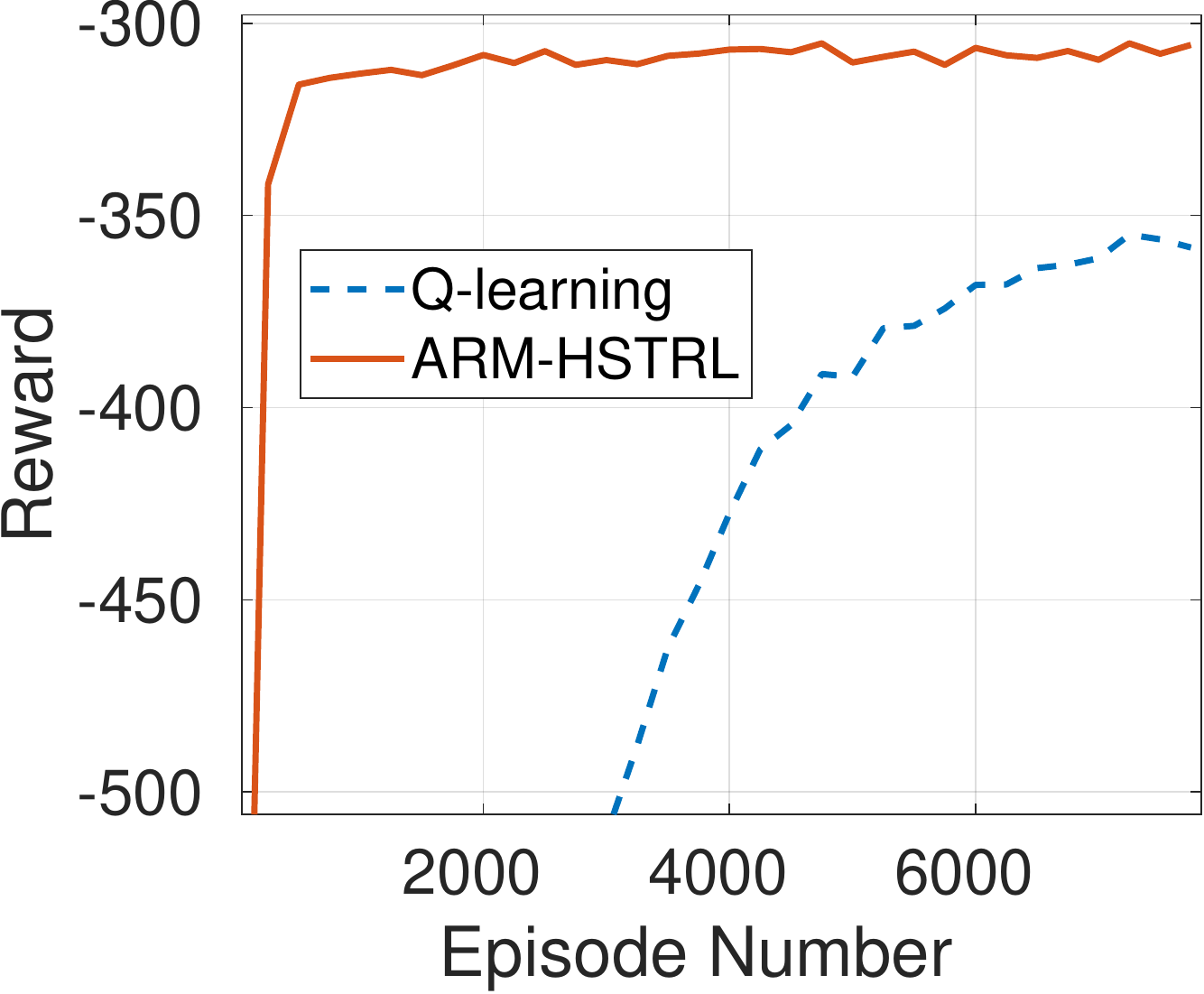}	
	\caption{ Comparison of receiving rewards along episodes. The comparison is between Q-learning and ARM-HSTRL in experiment 1, Figure \ref{fig:HST_test1_hi},  of the first testbed, Figure \ref{fig:testbed1}.}
	\label{fig:maze_reward_1_z}
\end{figure}

\begin{figure}	
	\centering
	\includegraphics[width=80mm, height=2in]{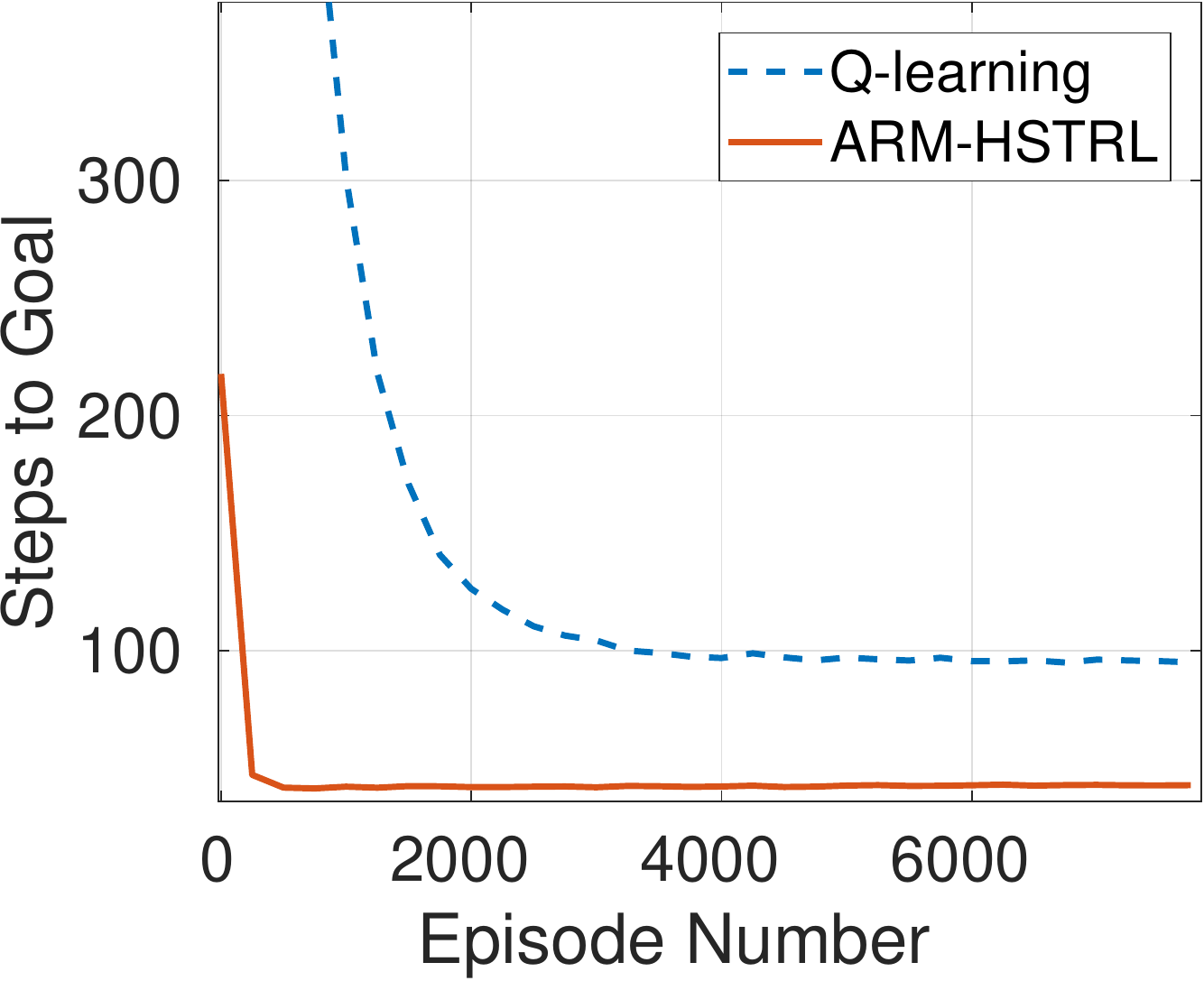}	
	\caption{ Represents the number of steps along episodes. The comparison is between Q-learning and ARM-HSTRL in experiment 2, Figure \ref{fig:HST_test2_hi},  of the first testbed, Figure \ref{fig:testbed1}.}
	\label{fig:action2zlabel}
\end{figure}

\begin{figure}	
	\centering
	\includegraphics[width=80mm, height=2in]{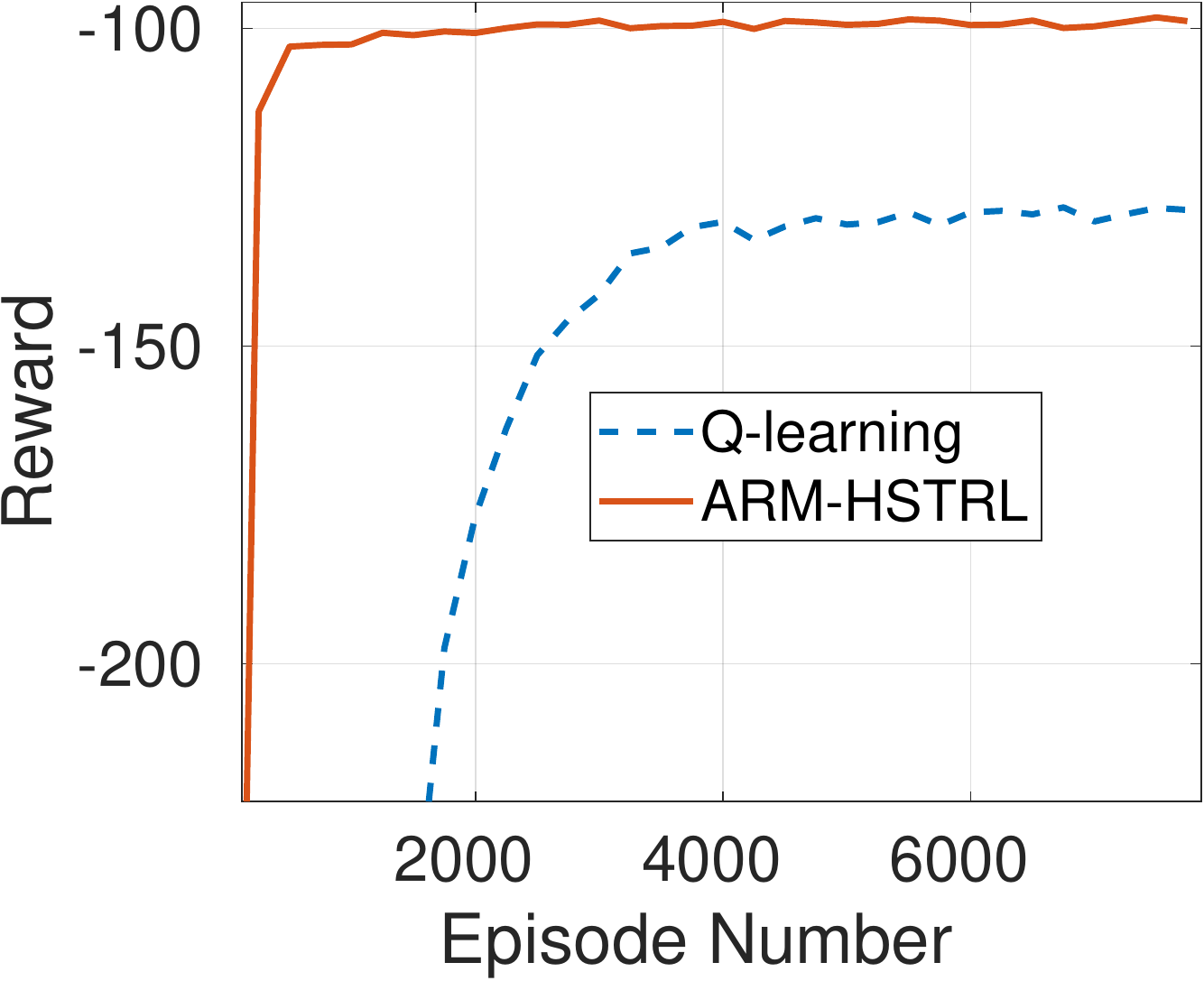}	
	\caption{ Comparison of receiving rewards along episodes. The comparison is between Q-learning and ARM-HSTRL in experiment 2, Figure \ref{fig:HST_test2_hi},  of the first testbed, Figure \ref{fig:testbed1}.}
	\label{fig:maze_reward_2_z}
\end{figure}

There is a significant difference in speed of learning between the proposed method and Q-learning in HRL in Figures \ref{fig:maze_reward_1_z} and \ref{fig:maze_reward_2_z}). The most important attribute of SMDP framework is using temporally extended actions to decrease the number of steps. It is shown in HRL in Figures \ref{fig:maze_action_1_z}  and \ref{fig:action2zlabel}, temporally extended actions considerably decrease the number of steps. p-values have been calculated between the proposed method and Q-learning in each diagram by using the t-test for $\alpha$ = $0.01$; the significant change is validated -- p-values are much smaller than 0.00001.

\textbf{MTRL}: In experiment 3, ARM-HSTRL is evaluated in MTRL on another maze, Figure \ref{fig:testbed_three}, for a complex the hierarchical structure, Figure \ref{fig:HST_test3}, as each of tasks has a different transition and a reward function. The task hierarchy has three different tasks as each of them has 5 sub-goals--it has $(676\times(5+1)\times3)=12681$ states. ARM-HSTRL is used to extract hierarchical structure of tasks, Figure \ref{fig:HST_test3}, in the testbed, Figure \ref{fig:testbed_three}, as each task has a different transaction and reward function. There are 2 runs and the results are averaged. In each run, there are 1000 episodes as in each episode a start state, a goal state, and a task is chosen randomly, the maximum number of actions for each episode is 10000. It means the agent should learn to find out which task is active now and then try to reach the goal state of that task. The value of \textit{minsup} is 0.3 and the value of \textit{minconf} is 0.6.

\begin{figure}	
	\centering
	\includegraphics[width=80mm, height=2in]{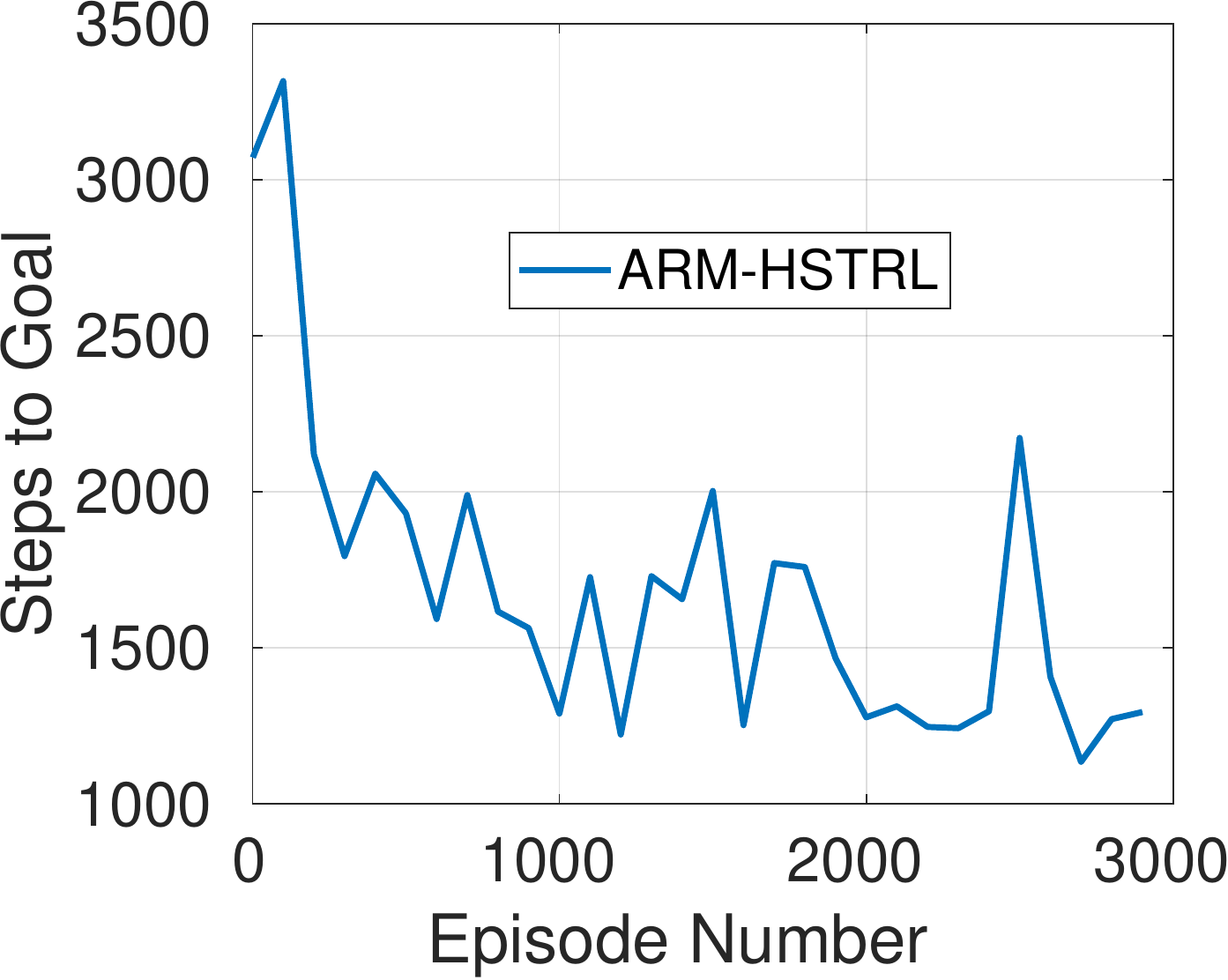}	
	\caption{ The number of steps of ARM-HSTRL along episode in experiment 3, Figure \ref{fig:HST_test3},  of the second testbed, Figure \ref{fig:testbed_three}.}
	\label{fig:maze_action_2}
\end{figure}

\begin{figure}	
	\centering
	\includegraphics[width=80mm, height=2in]{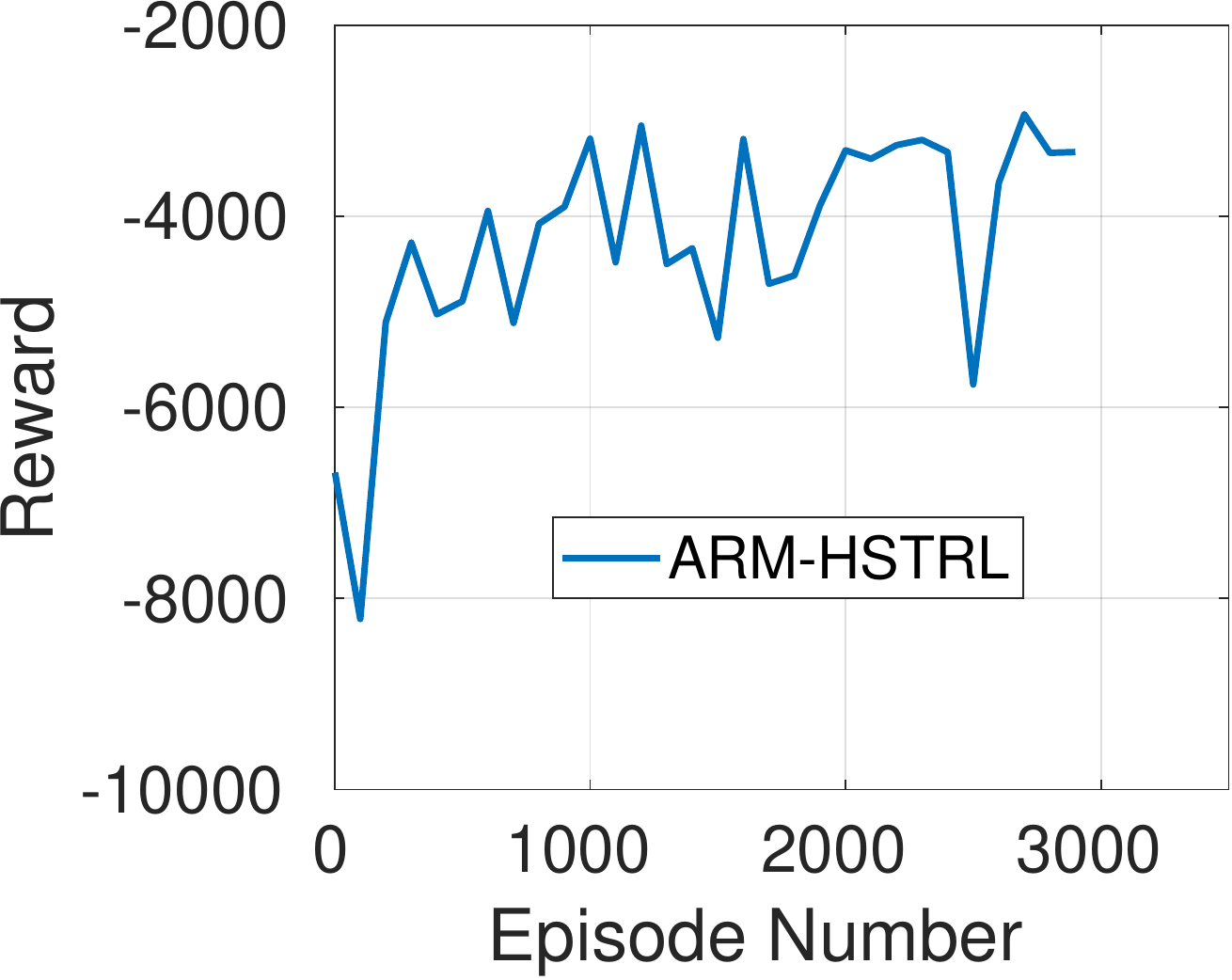}	
	\caption{ The amount of received rewards of ARM-HSTRL along episodes in experiment 3, Figure \ref{fig:HST_test3},  of the second testbed, Figure \ref{fig:testbed_three}.}
	\label{fig:maze_reward_2}
\end{figure}

Actions and their effects are same with the HRL part. One of the tasks, green or purple or yellow, is activated in each run randomly, the agent should enter in sub-goals states of activated task in the following order $1,2,3,4,5 $ and do the $press-key$ action in each of them, and then enters in the goal state of the run and do $press-key$ in that too, the agent receives a reward of $+10$, and the episode will be finished.  In MTRL, we just show  curves of ARM-HSTRL in Figure \ref{fig:maze_action_2}, and in Figure \ref{fig:maze_reward_2} since Q-learning cannot learn several tasks with different reward and transition functions.

\section{Conclusion}

\noindent ARM-HSTRL works based on trajectories of states and extracts sub-goals, as states which visited frequently in successful one, and relationships among them. The relationship among sub-goals of each task, decomposition of tasks, are grouped by each other in a hierarchical structure.  Specifically, two main parts of ARM, ``Frequent Itemset Generation'' and ``Rule Generation,'' are used to extract association rules. Each association rule alongside $t$ shows the sequence of sub-goals; thus, it shows the relation among them in a flat manner. In the second phase,  HST-construction, a main hierarchical structure of tasks is extracted by combining association rules. 

Unlike HI-MAT, the most recent work in task decomposition in HRL, ARM-HSTRL does not need the action model, not limited to factored MDP, can learn from different and several trajectories, and does not need to clean and to process the paths. Also, it is discussed and shown theoretically before, ARM-HSTRL is efficient, practical, and leads to hierarchical optimal policies. In addition, it is shown empirically in experimental results, the considerable improvement in the learning for two experiments in HRL.

The main contribution of the paper in  MTRL and TL is that when the hierarchical structure of tasks are extracted autonomously, it can distinguish and capture more tasks differences, and handle them in creating partial policies. For example, in experiment 3, each task has a different transition and a reward function. Also, the agent can distinguish tasks of each other, and separate or combine the learning of each task from other tasks depending on the similarity and difference among their structures. In other words, ARM-HSTRL can handle MTRL and transfer its learning among tasks with different transition functions. In fact, a lack of structural knowledge makes Q-learning impractical. 

The only work that used a task structure in TL \citep{mehta2008transfer} used a hand-made task structure to handle reward function differences.
ARM-HSTRL extracts the hierarchical structure of tasks autonomously, handles much more task differences like transition functions, and uses the structure of tasks to combine and separate learning of each task depending on the structure. To the best of our knowledge, no method has been proposed to provide the abilities in MTRL and TL. 
It is believed, the extracted hierarchical structure in the form of sub-tasks is supplementary, and a more robust and higher level of knowledge for transferring rather than sharing value functions. The effectiveness of transferring value functions is much more sensitive to kind and amount of similarity between source and target domains. Decomposed structure of tasks provides abstraction, and the agent can reuse, generalize, and transfer the knowledge for new domains.

\bibliographystyle{aaai} 
\bibliography{reff}

\end{document}